\documentclass[sigconf]{acmart}

\usepackage{amssymb}
\usepackage{amsmath}
\usepackage{color}
\usepackage{subcaption}
\usepackage{url}

\newcommand{\ignore}[1]{}

\begin{document}

\title[]{Honk: A PyTorch Reimplementation of Convolutional\\ Neural Networks for Keyword Spotting}

\author{Raphael Tang and Jimmy Lin}
\affiliation{\vspace{0.1cm}
  \department{David R. Cheriton School of Computer Science}
  \institution{University of Waterloo, Ontario, Canada}
}
\email{{r33tang, jimmylin}@uwaterloo.ca}

\begin{abstract}
We describe Honk, an open-source PyTorch reimplementation of
convolutional neural networks for keyword spotting that are included
as examples in TensorFlow. These models are useful for recognizing
``command triggers'' in speech-based interfaces (e.g., ``Hey Siri''),
which serve as explicit cues for audio recordings of utterances that
are sent to the cloud for full speech recognition. Evaluation on
Google's recently released Speech Commands Dataset shows that our
reimplementation is comparable in accuracy and provides a starting
point for future work on the keyword spotting task.
\end{abstract}

\maketitle

\section{Introduction}

Conversational agents that offer speech-based interfaces are
increasingly part of our daily lives, both embodied in mobile phones
as well as standalone consumer devices for the home. Prominent
examples include Google's Assistant, Apple's Siri, Amazon's Alexa, and
Microsoft's Cortana. Due to model complexity and computational
requirements, full speech recognition is typically performed in the
cloud:\ recorded audio is transferred to a datacenter for
processing. For both practical and privacy concerns, devices do not
continuously stream user speech into the cloud, but rely on a command
trigger, e.g., ``Hey Siri'', that provides an explicit cue signaling
input directed at the device. These verbal triggers also serve as an acknowledgment that
subsequent audio recordings of user utterances will be sent to backend servers and thus may be logged and analyzed.
A recent incident where user privacy expectations have not been met
involves the Google Home Mini smart speaker, when a reviewer discovered
that the device was surreptitiously recording his conversations without his knowledge or consent~\cite{spying}.
This incident demonstrates the importance of on-device
command triggering, which requires accurate,
low-powered keyword spotting capabilities.



Sainath and Parada~\cite{keywordcnn} proposed simple convolutional
neural network models for keyword spotting and reference
implementations are provided in TensorFlow.  These models, coupled with
the release of Google's Speech Commands Dataset~\cite{dataset}, provide a public
benchmark for the keyword spotting task.
This paper describes Honk, a PyTorch reimplementation of these models.
We are able to achieve recognition
accuracy comparable to the TensorFlow reference implementations.

\section{Data and Task}

A blog post from Google in August 2017~\cite{dataset} announced the
release of the Speech Commands Dataset, along with training and
inference code for convolutional neural networks for keyword spotting. 
The dataset, released
under a Creative Commons license, contains 65,000 one-second long
utterances of 30 short words by thousands of different
people. Additionally, the dataset contains such background noise samples as
pink noise, white noise, and human-made sounds. Quite explicitly, the
blog writes:

\begin{quote}
The dataset is designed to let you build basic but useful voice
interfaces for applications, with common words like ``Yes'', ``No'',
digits, and directions included.
\end{quote}

\noindent As such, this resource provides a nice benchmark for the
keyword spotting task that we are interested in.

Following Google's demo, our task is to discriminate among 10 of the 30 short
words, treating the rest of the 20 unused classes as a single ``unknown''
group of words. There is also one silence class comprised of soft
background noise to avoid misclassifying silence. Therefore, in total,
there are 12 output labels:\ ten keywords, one unknown class, and one
silence class.

\section{Implementation}

Honk, named after local fauna, is an open-source PyTorch
reimplementation of public TensorFlow keyword spotting
models,\footnote{\url{https://github.com/tensorflow/tensorflow/tree/master/tensorflow/examples/speech\_commands}}
which are in turn based on the work of Sainath and
Parada~\cite{keywordcnn}. In some cases, as we note below, details differ
between the two. We embarked on a PyTorch reimplementation
primarily to maintain consistency with the rest of our research
group's projects. However, we feel that
PyTorch has an advantage over TensorFlow in terms of readability
of the model specifications.

Following the TensorFlow reference, our implementation consists of two
distinct components:\ an input preprocessor and the convolutional neural
network models themselves. All our code is available on
GitHub\footnote{\url{https://github.com/castorini/honk}} for others to
build upon.

\subsection{Input Preprocessing} \label{sec:input_processing}

\begin{figure}
	\centering
	\includegraphics[width=0.42\textwidth]{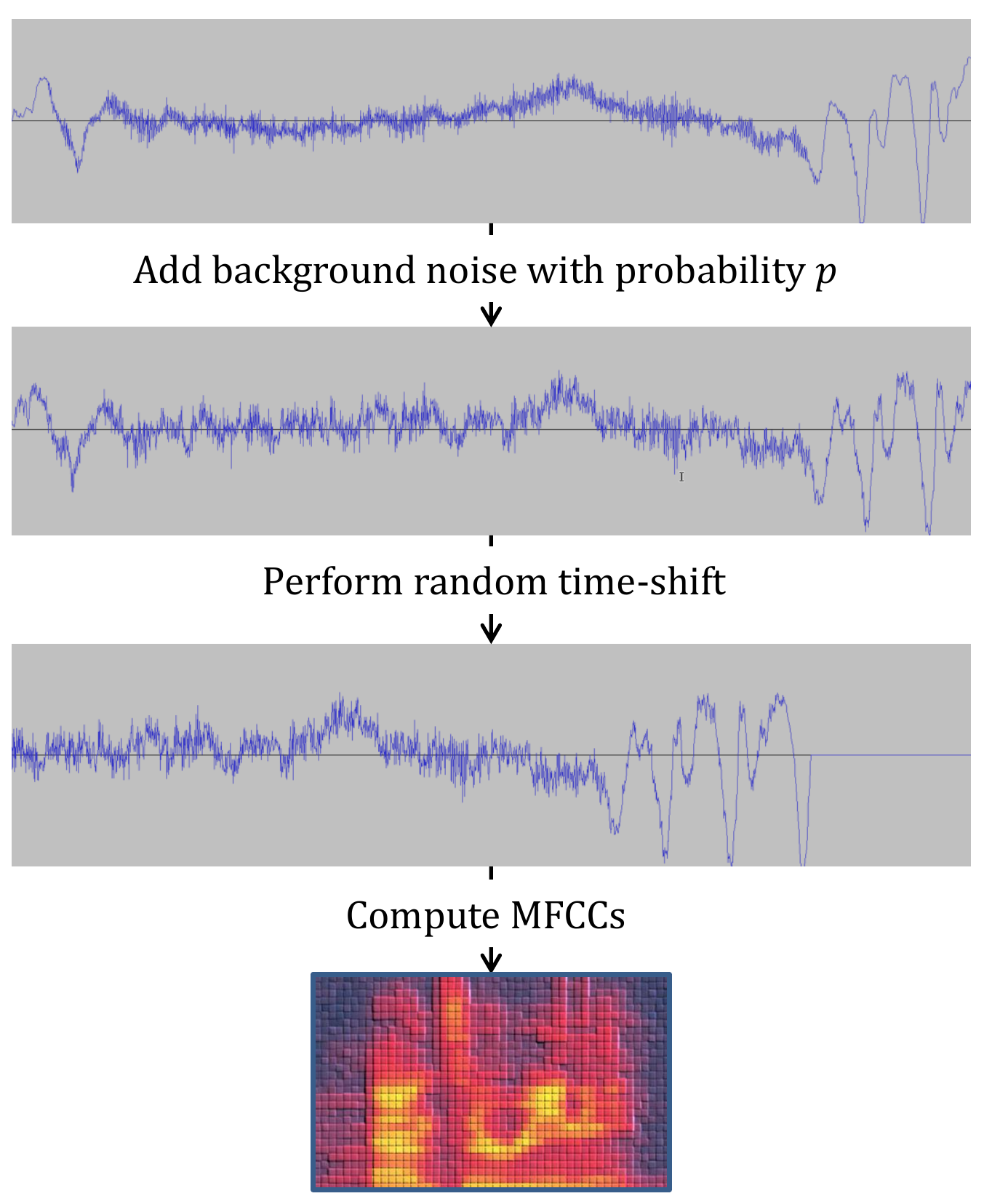}
	\caption{The input preprocessing pipeline.}
	\label{fig:input}
\end{figure}

Our PyTorch implementation uses the same preprocessing pipeline as the TensorFlow reference (see Figure~\ref{fig:input}).
To augment the dataset and to increase robustness, background noise consisting of 
white noise, pink noise, and human-made noise are mixed in with some of the 
input audio, and the sample is randomly time-shifted. For feature extraction, a 
band-pass filter of 20Hz/4kHz is first applied to reduce the effect of 
unimportant sounds. Forty-dimensional Mel-Frequency Cepstrum Coefficient (MFCC) 
frames are then constructed and stacked using a 30-millisecond window size and 
a 10-millisecond frame shift.

For the actual keyword spotting, Sainath and Parada \cite{keywordcnn} proposed to stack 23 frames to the left 
and 8 frames to the right at every frame for the input.
However, we followed the 
TensorFlow implementation and use as input the entire one-second stack. That 
is, our implementation stacks all 30-millisecond windows within the one-second 
sample, using a 10-millisecond frame shift.

\subsection{Model Architecture}

\begin{figure}
	\centering
	\includegraphics[width=0.45\textwidth]{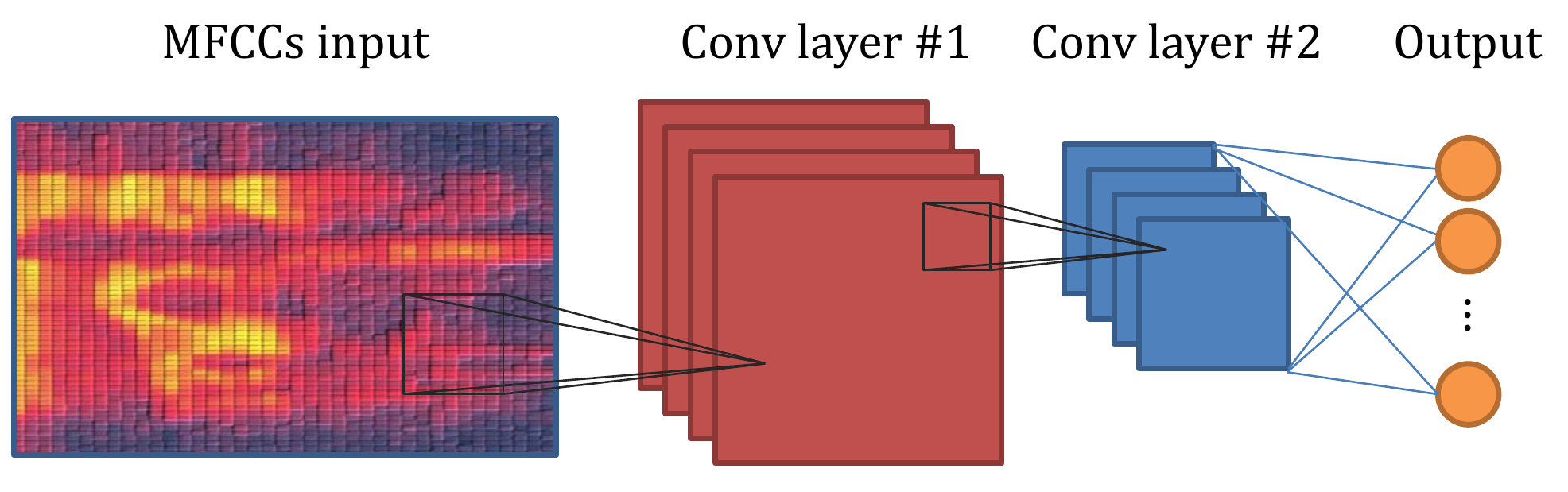}
	\caption{Convolutional neural network architecture for
		keyword spotting.}\label{fig:convnet}
\end{figure}

The basic model architecture for keyword spotting, shown in 
Figure~\ref{fig:convnet},
comprises one or more convolutional layers followed
by fully-connected hidden layers, ending with a softmax output. 
More specifically, an input of MFCCs $\mathbf{X} \in \mathbb{R}^{t\times f}$ is 
convolved with weights from the first convolutional layer, $\mathbf{W} \in
\mathbb{R}^{m \times r \times n}$, where $t$ and $f$ are the lengths 
in time and frequency, $m$ and $r$ are the width and height of the
convolution filter, and $n$ is the number of feature maps.
If desired, the convolution can stride by $s \times v$ and max-pool in $p 
\times q$, parameters which also influence the compactness of the model. 
Rectified linear units are used as the activation function for each 
non-linear layer.

Sainath and Parada~\cite{keywordcnn} proposed a model comprised of two
convolutional layers (as described above) with a linear layer, a
hidden layer, and a softmax layer for their full model, which they
referred to as \texttt{cnn-trad-fpool3}. They also devised compact
variants of their full model that reduce the number of parameters and
multiplies (for a low-power setting). We discuss our reimplementation
of the full model and its variants below.

\subsubsection{Full Model}

Our full model architecture is a faithful reimplementation of the full
TensorFlow model, which diverges slightly from the
\texttt{cnn-trad-fpool3} model in the Sainath and Parada paper. The
TensorFlow model makes a few changes, selecting $p = 2$ and $q = 2$
and dropping the hidden and linear layers in the original
paper. Surprisingly, we confirmed that this leads to better accuracy
for our task. We refer to this variant as
\texttt{cnn-trad-pool2}. For our task, with an input size of $101
\times 40$ and $n_{\text{labels}} = 12$, applying this architecture
(see details in Table \ref{table:arch-full}) results in $2.77\times10^7 +
7.08\times10^7 + 3.32\times10^5 = 9.88\times10^7$ multiply operations.
\begin{table}
	\begin{center}
		\begin{tabular}{ r | c @{\hspace{5mm}} c c c @{\hspace{5mm}} c
				@{\hspace{5mm}} c @{\hspace{5mm}} c} 
			\hline
			type & $m$ & $r$ & $n$ & $p$ & $q$ & $s$ & $v$ \\ 
			\hline
			conv & 20 & 8 & 64 & 2 & 2 & 1 & 1 \\
			conv & 10 & 4 & 64 & 1 & 1 & 1 & 1 \\
			softmax & - & - & $n_{\text{labels}}$ & - & - & - & - \\
			\hline
		\end{tabular}
	\end{center}
	\vspace{0.2cm}
	\caption{Parameters used in the \texttt{cnn-trad-pool2} model.}
	\label{table:arch-full}
\end{table}%
\begin{table}
	\begin{center}
		\begin{tabular}{ r | c @{\hspace{5mm}} c c c @{\hspace{5mm}} c  
				@{\hspace{5mm}} c @{\hspace{5mm}} c} 
			\hline
			type & $m$ & $r$ & $n$ & $p$ & $q$ & $s$ & $v$ \\ 
			\hline
			conv & $t$ & 8 & 186 & 1 & 1 & 1 & 1 \\
			hidden & - & - & 128 & - & - & - & -\\
			hidden & - & - & 128 & - & - & - & -\\
			softmax & - & - & $n_{\text{labels}}$ & - & - & - & - \\
			\hline		
		\end{tabular}
	\end{center}
	\vspace{0.2cm}
	\caption{Parameters used in TensorFlow's variant of 
		\texttt{cnn-one-fstride4}.}
	\label{table:arch-reduced}
\end{table}

\subsubsection{Compact Models}

Sainath and Parada~\cite{keywordcnn} proposed a few compact variants of their
full model that differ in pooling size
and the number of convolutional layers. Sacrificing some accuracy, these variants
have fewer parameters and multiply operations, specifically targeting 
low-powered devices.

We reimplemented all the variants but examined only TensorFlow's variant of
\texttt{cnn-one-fstride4} (see Table \ref{table:arch-reduced}), since it
obtains the best accuracy out of the compact variants that restrict
the number of multiplies performed. For our task, this architecture requires
$5.76\times10^6$ multiplies, more than an order of magnitude less than the 
number of multiplies in the full model. Note that only one convolutional layer is used for 
this model, and the TensorFlow variant performs no 
increased striding in frequency or time (see Table \ref{table:arch-reduced}). 

\section{Experimental Results}

For the purpose of attaining consistent comparisons, we closely replicate the 
same setup as in the TensorFlow reference implementation. Specifically, the task is to classify a 
short one-second utterance as ``yes'', ``no'', ``up'', ``down'', ``left'', 
``right'', ``on'', ``off'', ``stop'', ``go'', silence, or unknown.

Following the TensorFlow implementation, we initialized all biases to zero and all 
weights to samples from a truncated normal distribution with $\mu=0$ and 
$\sigma = 0.01$. We used stochastic gradient descent with a mini-batch size of 
100, learning rates of 0.001 and 0.01 for the full 
and compact models, respectively. We also ran our entire training/validation/test
process using five different random seeds, obtaining a distribution of the model
accuracy. For the full model, approximately 30 
epochs were required for convergence, while roughly 55 epochs were needed for the 
compact model.

Deviating from the TensorFlow implementation, we also tried training our models 
using stochastic gradient descent with a momentum of $0.9$. The compact model 
had failed to converge with a learning rate of 0.01, so the rate was decreased 
to 0.001. As shown in Table \ref{table:results}, we find that training with 
momentum yields improved results, especially for the full model.

The Speech Commands Dataset was split into training, validation, and test sets, with 80\%
in training, 10\% in validation, and 10\% in test. This results in roughly 
$22,000$ examples for training and $2,700$ each for validation and testing. Mirroring 
the TensorFlow implementation, for consistency across runs,
the hashed name of the audio file from the dataset determines 
which split the sample belongs to.
Specifically, the integer value of the SHA1 hash of the filename is used to 
place each example into either the training, validation, or test sets.

To generate training data via the process described in Section~\ref{sec:input_processing},
Honk adds background noise to each sample with a 
probability of $0.8$ at every epoch, where the noise is chosen randomly from the 
background noises provided in the Speech Commands Dataset. Our implementation 
also performs a random time-shift of $Y$ milliseconds before transforming the 
audio into MFCCs, where $Y\sim\textsc{Uniform}[-100, 100]$. In order to accelerate 
the training process, all preprocessed inputs are cached for reuse across different 
training epochs. At each epoch, 30\% of the cache is evicted.

\begin{table}
	\begin{center}
		\begin{tabular}{ r | c c} 
			\hline
			Model & Full & Compact \\
			\hline
			TensorFlow (TF)  & 87.8\% $\pm$ 0.435 & 77.4\% $\pm$ 0.839 \\
			PyTorch (PT)& 87.5\% $\pm$ 0.340 & 77.9\% $\pm$ 0.715 \\
			PT with momentum & 90.2\% $\pm$ 0.515 & 78.4\% $\pm$ 0.631\\
			\hline
		\end{tabular}
	\end{center}
        \vspace{0.2cm}
	\caption{Test accuracy along with 95\% confidence intervals
          for PyTorch and TensorFlow implementations of the full and compact models.}
	\label{table:results}
        \vspace{-0.3cm}
\end{table}%

We trained all our models using a workstation built from commodity hardware:\
dual GeForce GTX 1080 graphics cards, an i7-6800K CPU, and 64 GB
of RAM. This machine was more than sufficient to train the models
in this paper, all of which required less than 2 GB of GPU memory.

Our evaluation metric is accuracy, simply computed as the percentage of correct
forced choice predictions out of the examples in the test set.
Results are shown in Table \ref{table:results}, where we compare
the PyTorch and TensorFlow implementations of the full and compact models.
The reported accuracy is the mean
across all individual runs, accompanied by the 95\% confidence interval.
We find that the accuracies of the different implementations are comparable,
and the confidence intervals overlap. This suggests that we have
faithfully reproduced the TensorFlow models.

\section{Open-Source Codebase}

Beyond the implementation of the convolutional neural network models themselves
in our GitHub repository,\footnote{\url{https://github.com/castorini/honk}}
our codebase includes a number of additional features:

\begin{list}{\labelitemi}{\leftmargin=1.5em}

  \item A utility for recording and building custom speech commands,
    producing audio samples of the appropriate length and format.

  \item Test harnesses for training and testing any of a number of
    models implemented in TensorFlow and those proposed by Sainath and
    Parada~\cite{keywordcnn}

  \item A RESTful service that deploys a trained model. The server
    accepts base 64-encoded audio and responds with the predicted
    label of the utterance. This service can be used for on-device keyword
    spotting via local loopback.

  \item A desktop application for demonstrating the keyword spotting
    models described in this paper. The client uses the REST API above
    for model inference.

\end{list}

\noindent These features allow anyone to replicate the experiments
described in this paper, and provide a platform that others can build
on for the keyword spotting task.

\section{Conclusions and Future Work}

In this paper, we describe how two convolutional neural network models
from Sainath and Parada~\cite{keywordcnn} are implemented in practice
with TensorFlow, and we demonstrate that Honk is a faithful PyTorch
reimplementation of these models. As evaluated with the Google Speech
Commands Dataset, we find that the accuracies of the implementations
are comparable.

Directions for future work include exploring deployment on devices with limited 
computing power, applying different techniques to input data preprocessing, and 
developing a framework whereby command triggers can be easily added.

\section{Acknowledgments}

We would like to thank Zhucheng Tu for diligently reviewing Honk's source code.



\end{document}